\pdfoutput=1

\documentclass[11pt]{article}

\usepackage[preprint]{acl}

\usepackage{times}
\usepackage{latexsym}

\usepackage[T1]{fontenc}

\usepackage[utf8]{inputenc}

\usepackage{microtype}

\usepackage{inconsolata}

\usepackage{graphicx}

\usepackage{hyperref}       
\usepackage{url}            
\usepackage{booktabs}       
\usepackage{amsfonts}       
\usepackage{nicefrac}       
\usepackage{xcolor}         

\usepackage{amsmath}
\usepackage{makecell}
\usepackage{subcaption}
\usepackage{float}
\usepackage{multirow} 

\newcommand{\Scref}[1]{\S\ref{#1}}

\title{Probing LLM World Models: Enhancing Guesstimation with \\ Wisdom of Crowds Decoding}

\author{
    Yun-Shiuan Chuang \hspace{0.1cm}
    Sameer Narendran\textsuperscript{\textdagger} \hspace{0.1cm}
    Nikunj Harlalka\textsuperscript{
    \textdagger} \hspace{0.1cm}
    Alexander Cheung  \hspace{0.1cm} \\ 
    \textbf{Sizhe Gao}  \hspace{0.1cm}
    \textbf{Siddharth Suresh}  \hspace{0.1cm} 
    \textbf{Junjie Hu} \hspace{0.1cm} 
    \textbf{Timothy T. Rogers}\\
  University of Wisconsin-Madison\\
  \texttt{\{yunshiuan.chuang, nirunwiroj, snarendran, agoyal25\}@wisc.edu}\\
  \texttt{\{vfrigo, syang84, dshah, junjie.hu, ttrogers\}@wisc.edu}
}

\begin{document}

\maketitle
\def\thefootnote{\textdagger}\footnotetext{Joint second authors.}\def\thefootnote{\arabic{footnote}}

\begin{abstract}
Guesstimation—the task of making approximate quantitative estimates about objects or events—is a common real-world skill, yet remains underexplored in large language model (LLM) research. We introduce three guesstimation datasets: \textit{MARBLES}, \textit{FUTURE}, and \textit{ELECPRED}, spanning physical estimation (e.g., how many marbles fit in a cup) to abstract predictions (e.g., the 2024 U.S. presidential election). Inspired by the social science concept of \textit{Wisdom of Crowds} (WOC)—where the median of multiple estimates improves accuracy—we propose WOC decoding for LLMs. We replicate WOC effects in human participants and find that LLMs exhibit similar benefits: median aggregation across sampled responses consistently improves accuracy over greedy decoding, self-consistency decoding, and mean decoding. This suggests that LLMs encode a world model that supports approximate reasoning. Our results position guesstimation as a useful probe of LLM world knowledge and highlight WOC decoding as a strategy for enhancing LLM guesstimation performance on real-world tasks.

\end{abstract}

\section{Introduction}

Daily life often requires us to estimate uncertain quantities, from the crowd size at a political event to the weight of a turkey needed for a Thanksgiving dinner. In human populations, such ``guesstimation'' scenarios often exhibit {\it wisdom of crowds} (WOC) effects: in a random sample of estimates, the median lies closer to the ground truth than most individual guesses \citep{galton1907vox, yu2018literature}. WOC phenomena are thought to emerge from aggregating diverse individual \textit{world models}, each reflecting a person’s conceptual understanding of the world, which, when combined, can lead to surprisingly accurate estimates as individual errors cancel out. For instance, when estimating the number of jelly beans in a jar \citep{surowiecki2005wisdom}, people may rely on an implicit understanding of the typical size, shape, and firmness of jelly beans, and the shape, volume, and rigidity of the jar. Even for more abstract scenarios, people may also rely on general world-knowledge; for instance, when estimating the number of people requiring food stamps in Chicago, their guesses may reflect general knowledge/beliefs about poverty rates, accessibility of government programs, characteristics of large midwestern cities, etc. 

\begin{figure}[tb!] 
\centering
\includegraphics[width=1\linewidth]{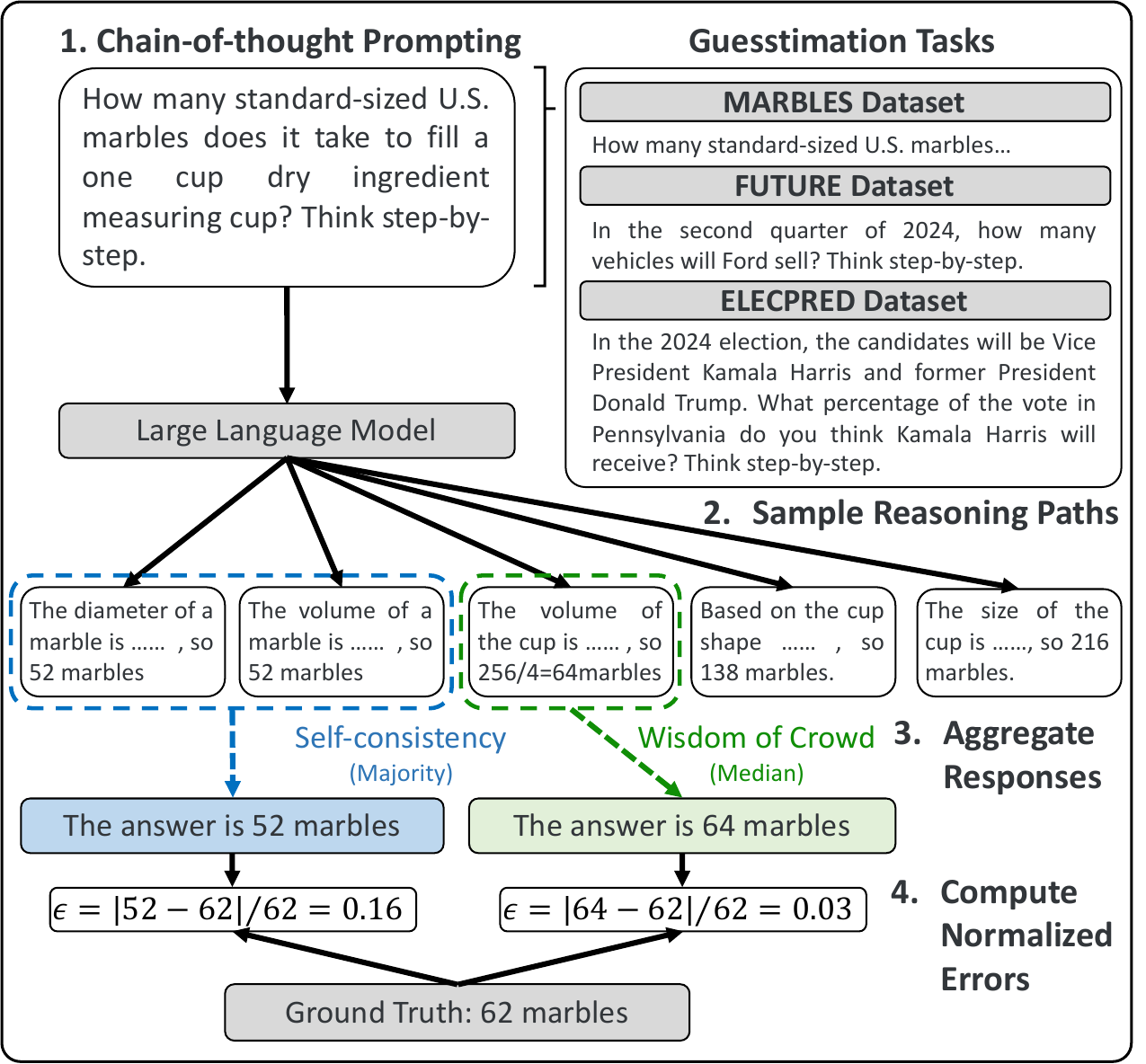}
\caption{LLM guesstimation through self-consistency decoding and wisdom of crowd (WOC) decoding.} 
\label{fig:diagram_prompts_three_tasks}
\end{figure}

\begin{table*}[!bht]
\centering
\Large
\resizebox{0.98\textwidth}{!}{
\begin{tabular}{l l l l l}
\toprule
\textbf{Model} & \textbf{Wisdom of Crowds} & \textbf{Self-Consistency} & \textbf{Mean Baseline} & \textbf{Greedy} \\
& (WOC; Median) & (Majority) & & \\
\midrule
Human Survey & \textbf{0.57} [0.54, 0.59] & 0.61 [0.57, 0.64] & 0.91 [0.80, 1.02] & -- \\
\cmidrule(lr){2-5}
Mistral & & & & \\
\hspace{1em} mistral-7b-instruct-v0.2 & \textbf{26.60} [21.39, 31.80] & 1154.61 [521.83, 1787.39] & 10004.69 [5196.11, 14813.27] & 1593.00 [487.33, 2698.67] \\
Mixtral & & & & \\
\hspace{1em} mixtral-8x7b-instruct-v0.1 & \textbf{1.57} [0.84, 2.30] & 28.11 [14.35, 41.87] & 80.40 [58.15, 102.65] & 12.81 [5.05, 20.58] \\
\hspace{1em} mixtral-8x22b-instruct-v0.1 & \textbf{1.33} [1.13, 1.54] & 33.66 [1.78, 65.54] & 55.73 [24.60, 86.86] & 4.79 [2.24, 7.34] \\
LLaMA 2 & & & & \\
\hspace{1em} llama-2-7b-chat-hf & \textbf{1.25} [0.89, 1.61] & 88.44 [1.12, 175.76] & 3704.85 [14.98, 7394.72] & 36.80 [7.32, 66.28] \\
\hspace{1em} llama-2-13b-chat-hf & \textbf{0.55} [0.47, 0.63] & 2.17 [1.17, 3.17] & 238.54 [26.56, 450.52] & 1.31 [0.84, 1.78] \\
\hspace{1em} llama-2-70b-chat-hf & \textbf{0.49} [0.38, 0.61] & 1.40 [0.68, 2.11] & 4555.78 [852.45, 8259.11] & 29.16 [13.08, 45.24] \\
LLaMA 3 & & & & \\
\hspace{1em} llama-3.1-8b-instruct & \textbf{0.81} [0.76, 0.85] & 0.94 [0.91, 0.97] & 4.80 [1.48, 8.12] & 2.80 [1.75, 3.85] \\
\hspace{1em} llama-3.1-70b-instruct & \textbf{0.49} [0.37, 0.61] & 1.07 [0.76, 1.39] & 3.12 [2.28, 3.96] & 6.55 [0.79, 12.30] \\
GPT & & & & \\
\hspace{1em} gpt-3.5-turbo-0125 & \textbf{0.64} [0.53, 0.74] & 0.73 [0.50, 0.95] & 1587.59 [10.4, 3164.78] & 16.82 [3.72, 29.93] \\
\hspace{1em} gpt-4-0125-preview & \textbf{1.00} [0.76, 1.23] & 1.07 [0.77, 1.37] & 1.29 [1.09, 1.49] & 1.04 [0.73, 1.34] \\
\bottomrule
\end{tabular}
}

\caption{Normalized errors ($\varepsilon$) across 30 sampled reasoning paths averaged over 15 guesstimation questions within the MARBLES dataset. The table is organized by model families and the four decoding strategies. Brackets denote standard errors. WOC is consistently the best decoding method. See Table~\ref{tab:results_future}, \ref{tab:results_elecpred} for results on the FUTURE and ELECPRED datasets.}
\label{tab:results_marbles}
\end{table*}



Here we assess whether contemporary large language models (LLMs) exhibit WOC phenomena similar to those observed in human populations. While a single LLM is not a true crowd, it is trained on vast amounts of linguistic data generated by many individual users, encoding a broad and diverse range of world knowledge. We interpret repeated prompting of an LLM as a way to elicit diverse probabilistic representations through stochastic sampling. This idea aligns with findings in cognitive science showing that repeated estimates from a single individual can produce a WOC effect, known as \textit{ the crowd within} \citep{vul2008measuring}. In this view, multiple samples from an LLM surface internal variability in its world model, similar to multiple estimates drawn from an individual’s internal probabilistic reasoning.


To systematically study guesstimation and WOC effects in LLMs, we created three guesstimation datasets: \textit{MARBLES}, \textit{FUTURE}, and \textit{ELECPRED}. The MARBLES dataset involves estimating the number of physical objects (e.g. marbles, coins) that can fit into different containers (e.g., one-cup dry-ingredients measuring cup), requiring reasoning based on real-world physical properties. On the other hand, FUTURE and ELECPRED datasets involve guesstimation in more abstract scenarios - predicting future real-world events like population growth, economic trends, or 2024 U.S. presidential election results, all of which require reasoning based on real-world knowledge such as demographics, economic conditions, and political trends.

The guesstimation questions were provided in natural language to the LLMs. To quantify the WOC effect in each case, we took the normalized error: the absolute difference between the median guess and the ground truth divided by the ground truth. The more these error terms are reduced with increasing crowd size, the greater the WOC advantage relative to an individual guesser. We further compared the LLM WOC behavior with the {\it self-consistency} decoding strategy, which samples model behavior many times and returns the majority vote among the samples, rather than the median as WOC. Prior work has suggested that self-consistency can improve model reasoning behavior \citep{wang2022self}.  In addition, we also conducted a human experiment and replicated previous findings about WOC in human crowds.

\begin{figure*}[tb!] 
\centering
\includegraphics[width=\linewidth]{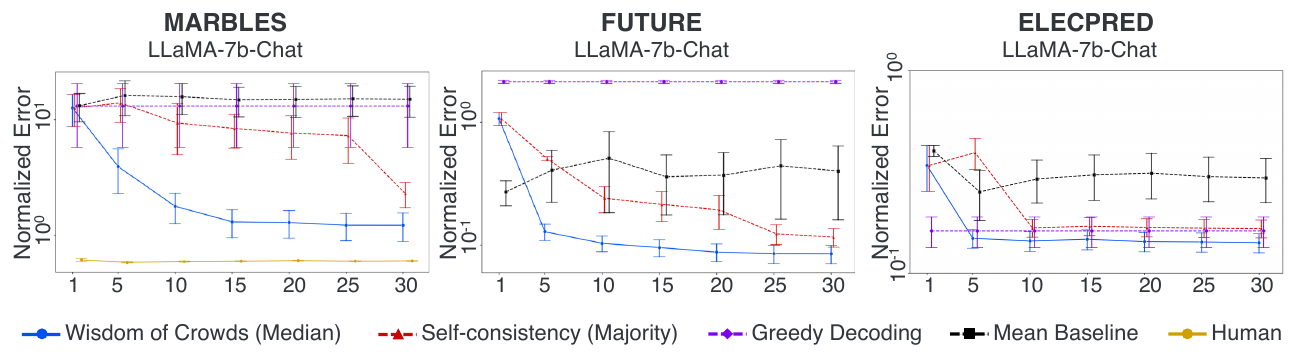}
\caption{Increased number of sampled reasoning paths boosts WOC accuracy, outperforming self-consistency, greedy decoding, and mean baseline across the three datasets. The normalized error averaged across all guesstimation questions within a dataset is shown on a logarithmic scale (y axis). The error bars denote standard errors.} 
\vspace{-10pt} 
\label{fig:line_plot_three_tasks}
\end{figure*}

Our results demonstrate the effectiveness of WOC decoding in guesstimation tasks in both humans and LLMs. We show that WOC decoding outperforms self-consistency, greedy decoding, and a mean baseline across both concrete and abstract guesstimation datasets and achieves greater accuracy with fewer samples. In sum, we propose guesstimation as a method to probe LLMs' world models, and showcase that we can apply WOC, a social science-inspired decoding strategy, to reach the best guesstimation performance.  Our findings have broader implications for real-world applications such as forecasting, which rely on an accurate world model. In sum, we introduce guesstimation as a new task that is very common in real world but has been overlooked by the NLP and AI community.

\section{Methods and Experimental Settings}
\label{sec:methods}

\paragraph{Guesstimation Datasets:} \textbf{1. MARBLES Dataset} consists of 15 guesstimation questions, involving five different containers (a one-cup dry ingredient measuring cup, a shot glass, a Starbucks iced tall cup, an Altoids tin, and a box for a deck of standard Bicycle playing cards) and three different items (standard-sized U.S. marbles, standard-sized M\&Ms, and U.S. quarters). For example, one question asks: \textit{``How many standard-sized U.S. marbles does it take to fill a one-cup dry ingredient measuring cup? Think step-by-step.''} The ground-truth answer for each question was determined by manually measuring each quantity three times and taking the median. To replicate previous findings about WOC in human crowds, we also conducted a human experiment (see Appendix \Scref{app:human_exp} for details). 






\textbf{2. FUTURE Dataset} consists of 15 guesstimation questions about predicting quantities of events in 2024, which was in the future at the time of model training but are now known. These quantities all come from a period after the pretraining cutoff date of the LLMs' training corpora, ensuring that the models could not rely on memorization but instead had to reason based on their world models. For example, one question asks: \textit{``In the second quarter of 2023, the number of vehicles Ford sold was 531,662. In the second quarter of 2024, how many vehicles will Ford sell? Think step-by-step.''} The pretraining cutoff dates of all LLMs we considered were before 2024.\footnote{The only exception was the \texttt{Mixtral-8x22b-instruct-v0.1} model, which has a cutoff date in Apr. 2024. Therefore, we excluded it when evaluating it on the FUTURE dataset.} The true answer for each question was determined based on information from credible websites (\Scref{app:list_questions}).

\textbf{3. ELECPRED Dataset} consists of 51 guesstimation questions, covering 50 U.S. states and Washington, D.C. The task required LLMs to predict the percentage of votes Kamala Harris would receive in the 2024 U.S. presidential election for each state. Similar to the \textit{FUTURE} dataset, the election occurred after all LLMs' pretraining cutoff dates. This ensured that the models could not rely on memorization but instead had to reason based on their world models about factors like demographics, historical trends, and political figures. The ground truth for each state was determined using official election results.

\paragraph{Large Language Models:} We tested the guesstimation capabilities in ten contemporary LLMs: five LLaMA models \citep{touvron2023llama}, a Mistral model \citep{jiang2023mistral}, two Mixtral models \citep{jiang2024mixtral}, and two GPT models. See \Scref{app:list_models} for the model details and \Scref{app:compute_resources} for compute resources.



\paragraph{Decoding Methods for Guesstimation:} For each guesstimation question, an LLM generates a response $x \in \mathbb{N}$, where there exists a ground truth $x^* \in \mathbb{N}$. We evaluate four decoding methods for LLM's guesstimation: \textit{wisdom of crowds} (WOC) decoding, \textit{self-consistency} decoding, \textit{greedy decoding} decoding, and a \textit{mean baseline} decoding. For the WOC and self-consistency methods, given a question, we sample $n$ reasoning paths (using chain-of-thought prompting; \citealp{wei2022chain, wei2022emergent}) from the LLM using temperature sampling with $T = 1$ (Figure~\ref{fig:diagram_prompts_three_tasks}). Each reasoning path yields a corresponding estimate $x$, resulting in a set of responses denoted as $\mathcal{X} = \{x_1, x_2, \dots, x_n\}$. For WOC, we take the median of the response set, $\text{median}(\mathcal{X}) = x_{\lceil \frac{n}{2} \rceil}$, as the final estimate. For self-consistency, we calculate the mode of the response set, $\text{mode} (\mathcal{X})$. In cases where the response set has multiple modes, we randomly choose one. For greedy decoding, the temperature is set to 0, making the response deterministic. For the mean baseline, we compute the arithmetic mean, $\text{mean} (\mathcal{X})$.

\paragraph{Evaluation Metric:} To assess the accuracy of the estimates across questions, we defined the normalized error. Formally, for a given estimate $\hat{x}$ and its corresponding ground truth $x^*$, the normalized error $\varepsilon$ is defined as: $\varepsilon = |\hat{x} - x^*|/x^*$. This metric is commonly used in literature on guesstimation tasks in human studies \citep{becker2017network,becker2019wisdom}.

\section{Results}
\label{sec:results}

\paragraph{Humans are Good at Guesstimation.} Human crowds achieve highly accurate guesstimation under WOC decoding ($\varepsilon=0.57$) compared to most LLMs in the MARBLES dataset (Table~\ref{tab:results_marbles}). This replicates previous findings about WOC in humans \citep{galton1907vox, yu2018literature}. In addition, WOC decoding has a higher accuracy compared to self-consistency decoding ($\varepsilon: 0.57 < 0.61$) and the mean baseline ($\varepsilon = 0.91$).


\paragraph{Wisdom of Crowds (WOC) Decoding Supports Guesstimation in LLMs.} For LLMs, the WOC decoding method consistently outperforms the self-consistency, greedy decoding, and the mean baseline in the three guesstimation datasets and across different model variants (Table~\ref{tab:results_marbles}, \ref{tab:results_future}, \ref{tab:results_elecpred}). In a few cases, other decoding strategies achieves the same accuracy as WOC decoding, but WOC is consistently among the best decoding methods. We conducted statistical tests and showed that the superiority of WOC decoding is statistically significant across LLMs and datasets (see \Scref{app:stat_tests}).

\paragraph{WOC Performance Improves More Efficiently than Self-Consistency.} Increasing the number of sampled reasoning paths improves the accuracy of the WOC decoding method (Figure~\ref{fig:line_plot_three_tasks}). In contrast, while increasing the sample size also leads to better guesstimation performance of the self-consistency method, the improvement is much slower than the WOC decoding method. For example, for FUTURE and ELECPRED datasets, WOC decoding using 5 samples achieves higher accuracy than self-consistency decoding using 30 samples.


\paragraph{WOC Decoding Yields the Most Accurate Election Forecast.} WOC decoding outperforms self-consistency and greedy decoding in predicting Kamala Harris’s 2024 vote share across U.S. states (Table~\ref{tab:results_elecpred}). To better interpret these predictions, we convert state-level vote shares into electoral votes and visualized the results on a national map (see \S\ref{app:elecpred_map}). While WOC decoding achieves the most accurate prediction, it shows an overall bias favoring Democrats. Understanding the source of this bias remains an open question for future research.

\paragraph{Distributional Robustness of WOC Decoding.} We investigated whether WOC decoding’s effectiveness stems from specific properties of the response distribution (e.g., skewness or variance). Our analysis found no consistent correlation between these properties and WOC gains, suggesting its robustness is not distribution-dependent. This is in line with human WOC literature \citep{galton1907vox, surowiecki2005wisdom}. Full results are in \Scref{app:dist_analysis}. “In addition, we present qualitative examples showing how WOC decoding can outperform self-consistency, greedy decoding, and the mean baseline (\Scref{app:qual_examples}).




\section{Related Work}
\paragraph{Guesstimation and Wisdom of Crowds (WOC).} For a crowd to reach better guesstimation, WOC has proven to be effective, as long as individual estimates within the group are statistically independent \citep{surowiecki2005wisdom,nofer2015crowds}. This independence ensures that their errors are uncorrelated, allowing them to cancel out in aggregate. Among aggregation strategies, taking the median has been shown to be robust regardless of the response distributional properties \citep{hora2013median, davis2014crowd}. WOC has demonstrated practical success in domains such as market prediction and political forecasting \citep{yu2018literature}.




\paragraph{Prompting and Decoding Strategies for LLM Reasoning.} Prompting methods aim to guide LLMs in generating useful outputs, particularly for reasoning tasks. Chain-of-thought (CoT) prompting has been shown to improve performance by encouraging intermediate reasoning steps, both in few-shot and zero-shot settings \citep{wei2022chain, kojima2022large}. However, the variability in generated CoT responses has led to the development of more robust decoding strategies. One such method is self-consistency decoding, which samples multiple reasoning paths and selects the most frequent answer \citep{wang2022self}. While effective in some cases, later work found it to be unreliable in others \citep{nguyen2024consistent, byerly2024effective}. To our knowledge, we are the first to apply the WOC decoding strategy to LLM reasoning.

\section{Conclusion}
\label{sec:conclusion}

In this study, we show that LLMs possess a world model necessary for effective guesstimation, a common yet overlooked task in the AI community. To evaluate this, we introduce three guesstimation datasets: \textit{MARBLES}, \textit{FUTURE}, and \textit{ELECPRED}, where one must estimate both concrete and abstract quantities based on knowledge about the world. Similar to humans, LLMs also exhibit the WOC effect, in which the median of estimates leads to more accurate results than greedy decoding, self-consistency, and the mean baseline. In addition, WOC performance improves more efficiently than self-consistency as the number of sampled reasoning paths increases. In sum, we introduce guesstimation as a new task that is very common in the real world yet has been largely overlooked by the NLP and AI community.

\section*{Limitations}

\paragraph{The Scope of Guesstimation Questions is U.S.-Centric} Our guesstimation questions are heavily U.S.-centric, covering topics such as common U.S. household items, U.S. economic statistics, and U.S. election results. It remains unclear whether LLMs would perform equally well on guesstimation tasks in other cultural and geographical contexts. Prior work suggests that LLMs tend to perform better on U.S.-centric tasks due to imbalanced training data \citep{chu2024fairness}. To explore this limitation, we conducted a supplementary experiment using the 2025 German Federal Election, following the same format as the ELECPRED dataset. We found that LLM performance in this context was weaker and the benefit of WOC decoding less consistent. These findings align with previous work and highlight the need to examine generalizability across countries and cultures. However, such cross-cultural evaluation is beyond the scope of this paper and we leave a more systematic investigation to future work. Full details and results are provided in \Scref{app:result_german_elecpred}.

\paragraph{Mechanism Behind WOC’s Superiority} While we find that WOC decoding consistently outperforms self-consistency, the underlying mechanism driving this improvement remains unclear. One hypothesis is that taking the median mitigates the influence of extreme outlier predictions, making WOC more robust. To test potential explanations, we conducted a distributional analysis of the model responses (e.g., skewness, variance) and found no consistent correlation between these properties and WOC gains, suggesting that WOC’s effectiveness is not simply due to favorable distributional characteristics (see \S\ref{app:dist_analysis}). While this rules out some surface-level statistical explanations, further work is needed to understand the deeper cognitive or algorithmic mechanisms behind WOC’s advantage and whether they generalize across other reasoning tasks.

\paragraph{Dataset Reusability}
Two of our datasets—\textit{FUTURE} and \textit{ELECPRED}—contain questions about real-world events that occurred after 2024. As such, they are best suited for evaluating LLMs with training cutoff dates before 2024. While future LLMs with later cutoffs may still be evaluated on these datasets, there is a risk of data contamination if relevant information has appeared in their training corpus. This limitation is not unique to our work and also applies to many widely used benchmarks.

Importantly, our contribution lies not only in releasing specific datasets but in introducing a generalizable methodology for constructing guesstimation benchmarks. The \textit{FUTURE} dataset, for example, can be dynamically updated with new questions involving future events (e.g., forecasting vehicle sales in Q2 2025) as soon as ground-truth data becomes available. This evolving design enables the task to remain temporally forward-looking and continues to serve as a diagnostic tool for evaluating LLMs’ world models.

\section*{Ethics Statement}

For the human experiment, our study has been reviewed and approved by the Institutional Review Board (IRB) of our institution. In addition, we will release our code base solely for research purposes, and adhere to the terms of use by OpenAI's API \footnote{\url{https://openai.com/policies/terms-of-use}} and their MIT license \footnote{\url{https://github.com/openai/openai-openapi/blob/master/LICENSE}}, as well as Mistral AI's non-production license (MNPL) \footnote{\url{https://mistral.ai/licenses/MNPL-0.1.md}} and Meta's Llama community license \footnote{https://www.llama.com/faq/}.

\section*{Acknowledgements}

We thank the reviewers, the area chair for their feedback. This work was funded by the Multi University Research Initiative grant from the Department of Defense, W911NF2110317 (with Rogers as Co-I) and a Research Forward award from the University of Wisconsin-Madison (Rogers, PI).


\newpage
\bibliography{custom}

\begin{thebibliography}{21}
\providecommand{\natexlab}[1]{#1}

\bibitem[{Becker et~al.(2017)Becker, Brackbill, and Centola}]{becker2017network}
Joshua Becker, Devon Brackbill, and Damon Centola. 2017.
\newblock Network dynamics of social influence in the wisdom of crowds.
\newblock \emph{Proceedings of the National Academy of Sciences of the United States of America}, 114(26):E5070.

\bibitem[{Becker et~al.(2019)Becker, Porter, and Centola}]{becker2019wisdom}
Joshua Becker, Ethan Porter, and Damon Centola. 2019.
\newblock The wisdom of partisan crowds.
\newblock \emph{Proceedings of the National Academy of Sciences of the United States of America}, 116(22):10717--10722.

\bibitem[{Byerly and Khashabi(2024)}]{byerly2024effective}
Adam Byerly and Daniel Khashabi. 2024.
\newblock How effective is self-consistency for long-context problems?
\newblock \emph{arXiv preprint arXiv:2411.01101}.

\bibitem[{Chu et~al.(2024)Chu, Wang, and Zhang}]{chu2024fairness}
Zhibo Chu, Zichong Wang, and Wenbin Zhang. 2024.
\newblock Fairness in large language models: A taxonomic survey.
\newblock \emph{ACM SIGKDD explorations newsletter}, 26(1):34--48.

\bibitem[{Davis-Stober et~al.(2014)Davis-Stober, Budescu, Dana, and Broomell}]{davis2014crowd}
Clintin~P Davis-Stober, David~V Budescu, Jason Dana, and Stephen~B Broomell. 2014.
\newblock When is a crowd wise?
\newblock \emph{Decision}, 1(2):79.

\bibitem[{Galton(1907)}]{galton1907vox}
Francis Galton. 1907.
\newblock Vox populi.
\newblock \emph{Nature}, 75(1949):450--451.

\bibitem[{{History, Art \& Archives, U.S. House of Representatives}()}]{houseofrepselecresults}
{History, Art \& Archives, U.S. House of Representatives}.
\newblock \href {https://history.house.gov/Institution/Election-Statistics/} {{Election Statistics: 1920 to Present}}.
\newblock Accessed: February 11, 2025.

\bibitem[{Hora et~al.(2013)Hora, Fransen, Hawkins, and Susel}]{hora2013median}
Stephen~C Hora, Benjamin~R Fransen, Natasha Hawkins, and Irving Susel. 2013.
\newblock Median aggregation of distribution functions.
\newblock \emph{Decision Analysis}, 10(4):279--291.

\bibitem[{Jiang et~al.(2023)Jiang, Sablayrolles, Mensch, Bamford, Chaplot, Casas, Bressand, Lengyel, Lample, Saulnier et~al.}]{jiang2023mistral}
Albert~Q Jiang, Alexandre Sablayrolles, Arthur Mensch, Chris Bamford, Devendra~Singh Chaplot, Diego de~las Casas, Florian Bressand, Gianna Lengyel, Guillaume Lample, Lucile Saulnier, et~al. 2023.
\newblock Mistral 7b.
\newblock \emph{arXiv preprint arXiv:2310.06825}.

\bibitem[{Jiang et~al.(2024)Jiang, Sablayrolles, Roux, Mensch, Savary, Bamford, Chaplot, Casas, Hanna, Bressand et~al.}]{jiang2024mixtral}
Albert~Q Jiang, Alexandre Sablayrolles, Antoine Roux, Arthur Mensch, Blanche Savary, Chris Bamford, Devendra~Singh Chaplot, Diego de~las Casas, Emma~Bou Hanna, Florian Bressand, et~al. 2024.
\newblock Mixtral of experts.
\newblock \emph{arXiv preprint arXiv:2401.04088}.

\bibitem[{Kojima et~al.(2022)Kojima, Gu, Reid, Matsuo, and Iwasawa}]{kojima2022large}
Takeru Kojima, Shixiang Gu, Mike Reid, Yutaka Matsuo, and Kazuto Iwasawa. 2022.
\newblock Large language models are zero-shot reasoners.
\newblock \emph{arXiv preprint arXiv:2205.11916}.

\bibitem[{Nguyen et~al.(2024)Nguyen, Mekala, Dong, and Shang}]{nguyen2024consistent}
Alex Nguyen, Dheeraj Mekala, Chengyu Dong, and Jingbo Shang. 2024.
\newblock When is the consistent prediction likely to be a correct prediction?
\newblock \emph{arXiv preprint arXiv:2407.05778}.

\bibitem[{Nofer and Nofer(2015)}]{nofer2015crowds}
Michael Nofer and Michael Nofer. 2015.
\newblock Are crowds on the internet wiser than experts?--the case of a stock prediction community.
\newblock \emph{The Value of Social Media for Predicting Stock Returns: Preconditions, Instruments and Performance Analysis}, pages 27--61.

\bibitem[{Surowiecki(2005)}]{surowiecki2005wisdom}
James Surowiecki. 2005.
\newblock \emph{The wisdom of crowds}.
\newblock Anchor.

\bibitem[{Touvron et~al.(2023)Touvron, Martin, Stone, Albert, Almahairi, Babaei, Bashlykov, Batra, Bhargava, Bhosale et~al.}]{touvron2023llama}
Hugo Touvron, Louis Martin, Kevin Stone, Peter Albert, Amjad Almahairi, Yasmine Babaei, Nikolay Bashlykov, Soumya Batra, Prajjwal Bhargava, Shruti Bhosale, et~al. 2023.
\newblock Llama 2: Open foundation and fine-tuned chat models.
\newblock \emph{arXiv preprint arXiv:2307.09288}.

\bibitem[{Vul and Pashler(2008)}]{vul2008measuring}
Edward Vul and Harold Pashler. 2008.
\newblock Measuring the crowd within: Probabilistic representations within individuals.
\newblock \emph{Psychological Science}, 19(7):645--647.

\bibitem[{Wang et~al.(2023)Wang, Wei, Schuurmans, Le, Chi, Narang, Chowdhery, and Zhou}]{wang2022self}
Xuezhi Wang, Jason Wei, Dale Schuurmans, Quoc~V Le, Ed~H Chi, Sharan Narang, Aakanksha Chowdhery, and Denny Zhou. 2023.
\newblock Self-consistency improves chain of thought reasoning in language models.
\newblock In \emph{The Eleventh International Conference on Learning Representations}.

\bibitem[{Wei et~al.(2022{\natexlab{a}})Wei, Tay, Bommasani, Raffel, Zoph, Borgeaud, Yogatama, Bosma, Zhou, Metzler, Chi, Hashimoto, Vinyals, Liang, Dean, and Fedus}]{wei2022emergent}
Jason Wei, Yi~Tay, Rishi Bommasani, Colin Raffel, Barret Zoph, Sebastian Borgeaud, Dani Yogatama, Maarten Bosma, Denny Zhou, Donald Metzler, Ed~H. Chi, Tatsunori Hashimoto, Oriol Vinyals, Percy Liang, Jeff Dean, and William Fedus. 2022{\natexlab{a}}.
\newblock \href {https://openreview.net/forum?id=yzkSU5zdwD} {Emergent abilities of large language models}.
\newblock \emph{Transactions on Machine Learning Research}.
\newblock Survey Certification.

\bibitem[{Wei et~al.(2022{\natexlab{b}})Wei, Wang, Schuurmans, Bosma, Xia, Chi, Le, Zhou et~al.}]{wei2022chain}
Jason Wei, Xuezhi Wang, Dale Schuurmans, Maarten Bosma, Fei Xia, Ed~Chi, Quoc~V Le, Denny Zhou, et~al. 2022{\natexlab{b}}.
\newblock Chain-of-thought prompting elicits reasoning in large language models.
\newblock \emph{Advances in Neural Information Processing Systems}, 35:24824--24837.

\bibitem[{Wilcoxon(1945)}]{wilcoxon1945}
Frank Wilcoxon. 1945.
\newblock \href {https://doi.org/10.2307/3001968} {Individual comparisons by ranking methods}.
\newblock \emph{Biometrics Bulletin}, 1(6):80--83.

\bibitem[{Yu et~al.(2018)Yu, Chai, and Liu}]{yu2018literature}
Chao Yu, Yueting Chai, and Yi~Liu. 2018.
\newblock Literature review on collective intelligence: a crowd science perspective.
\newblock \emph{International Journal of Crowd Science}, 2(1):64--73.

\end{thebibliography}

\newpage
\appendix

\newpage
\appendix
\label{sec:appendix}

\section{Results on FUTURE and ELECPRED Datasets}
\label{app:results_future_elecpred}
The results for the \textit{FUTURE} and \textit{ELECPRED} datasets (Tables~\ref{tab:results_future} and \ref{tab:results_elecpred}) mirror the findings observed in the \textit{MARBLES} dataset. Across all three datasets, WOC decoding is consistently the best decoding strategy.

\begin{table*}[!bht]
\centering
\Large
\resizebox{0.98\textwidth}{!}{
\begin{tabular}{l l l l l}
\toprule
\textbf{Model} & \textbf{Wisdom of Crowds} & \textbf{Self-Consistency} & \textbf{Mean Baseline} & \textbf{Greedy} \\
& (WOC; Median) & (Majority) & & \\
\midrule
  Mistral & & & \\
  \hspace{1em} mistral-7b-instruct-v0.2 & \textbf{0.61} [0.47, 0.75] & 0.91 [0.84, 0.97] & 0.58 [0.36, 0.80] & 1.79 [0.38, 3.20] \\
  Mixtral & & & \\           
  \hspace{1em} mixtral-8x7b-instruct-v0.1 & \textbf{0.09} [0.06, 0.12] & \textbf{0.09} [0.06, 0.11] & \textbf{0.09} [0.07, 0.11] & 0.60 [0.16, 1.04] \\    
  LLaMA 2 & & & \\                  
  \hspace{1em} llama-2-7b-chat-hf & \textbf{0.08} [0.06, 0.11] & 1.19 [0.19, 2.18] & 0.47 [0.24, 0.70] & 2.45 [1.00, 3.89] \\ 
  \hspace{1em} llama-2-13b-chat-hf & \textbf{0.09} [0.05, 0.12] & 7.53 [1.27, 13.80] & 4635.86 [11.95, 9259.77] & 0.11 [0.07, 0.15] \\ 
  \hspace{1em} llama-2-70b-chat-hf & \textbf{0.09} [0.06, 0.11] & 4.57 [0.41, 8.73] & 0.11 [0.08, 0.14] & 0.19 [0.11, 0.28] \\ 
  LLaMA 3 & & & \\             
  \hspace{1em} llama-3.1-8b-instruct & \textbf{0.54} [0.42, 0.65] & 7.84 [1.60, 14.08] & 13339.02 [2235.59, 24442.45] & 8.54 [2.20, 14.89] \\ 
  \hspace{1em} llama-3.1-70b-instruct & \textbf{0.09} [0.06, 0.12] & 0.10 [0.07, 0.12] & 0.30 [0.10, 0.50] & 0.10 [0.07, 0.13] \\ 
  GPT & & & \\         
  \hspace{1em} gpt-3.5-turbo-0125 & \textbf{0.10} [0.06, 0.13] & \textbf{0.10} [0.06, 0.13] & \textbf{0.10} [0.07, 0.13] & \textbf{0.10} [0.06, 0.13] \\ 
  \hspace{1em} gpt-4-0125-preview & \textbf{0.08} [0.06, 0.11] & 0.09 [0.07, 0.12] & 0.16 [0.12, 0.20] & \textbf{0.08} [0.06, 0.11] \\ 

\bottomrule
\end{tabular}
}
\caption{Normalized errors ($\varepsilon$) across 30 sampled reasoning paths for each LLM on the FUTURE dataset. The table is organized by model families and shows results under three decoding strategies. Brackets denote standard errors based on 30 bootstrapped samples. WOC is consistently the best decoding method.}
\label{tab:results_future}
\end{table*}

\begin{table*}[!bht]
\centering
\small
\resizebox{0.98\textwidth}{!}{
\begin{tabular}{l l l l l}
\toprule
\textbf{Model} & \textbf{Wisdom of Crowds} & \textbf{Self-Consistency} & \textbf{Mean Baseline} & \textbf{Greedy} \\
& (WOC; Median) & (Majority) & & \\
\midrule
  Mistral & & & \\
  \hspace{1em} mistral-7b-instruct-v0.2 & \textbf{0.07} [0.06, 0.07] & 0.11 [0.10, 0.13] & \textbf{0.07} [0.06, 0.08] & 0.16 [0.13, 0.20] \\
  Mixtral & & & \\           
  \hspace{1em} mixtral-8x7b-instruct-v0.1 & \textbf{0.05} [0.05, 0.06] & 0.06 [0.06, 0.07] & \textbf{0.05} [0.05, 0.05] & 0.09 [0.07, 0.11] \\    
  \hspace{1em} mixtral-8x22b-instruct-v0.1 & \textbf{0.06} [0.05, 0.07] & \textbf{0.06} [0.06, 0.07] & \textbf{0.06} [0.06, 0.06] & 0.12 [0.10, 0.13] \\           
  LLaMA 2 & & & \\                  
  \hspace{1em} llama-2-7b-chat-hf & \textbf{0.14} [0.12, 0.16] & 0.16 [0.15, 0.18] & 0.30 [0.22, 0.38] & 0.16 [0.13, 0.19] \\ 
  \hspace{1em} llama-2-13b-chat-hf & \textbf{0.10} [0.09, 0.11] & 0.12 [0.11, 0.13] & \textbf{0.10} [0.09, 0.11] & 0.16 [0.12, 0.19] \\ 
  \hspace{1em} llama-2-70b-chat-hf & 0.11 [0.09, 0.12] & 0.12 [0.11, 0.14] & \textbf{0.10} [0.09, 0.11] & 0.12 [0.11, 0.13] \\ 
  LLaMA 3 & & & \\             
  \hspace{1em} llama-3.1-8b-instruct & 0.07 [0.06, 0.07] & 0.08 [0.07, 0.09] & \textbf{0.06} [0.06, 0.06] & 0.08 [0.07, 0.08] \\ 
  \hspace{1em} llama-3.1-70b-instruct & \textbf{0.05} [0.05, 0.06] & \textbf{0.05} [0.05, 0.06] & \textbf{0.05} [0.05, 0.05] & 0.08 [0.06, 0.10] \\ 
  GPT & & & \\         
  \hspace{1em} gpt-3.5-turbo-0125 & 0.07 [0.06, 0.07] & 0.08 [0.07, 0.08] & \textbf{0.06} [0.05, 0.07] & 0.16 [0.12, 0.20] \\ 
  \hspace{1em} gpt-4-0125-preview & \textbf{0.05} [0.05, 0.06] & \textbf{0.05} [0.04, 0.05] & 0.09 [0.09, 0.09] & \textbf{0.05} [0.05, 0.06] \\ 

\bottomrule
\end{tabular}
}
\caption{Normalized errors ($\varepsilon$) across 30 sampled reasoning paths for each LLM on the ELECPRED dataset. The table is organized by model families and shows results under three decoding strategies. Brackets denote standard errors based on 30 bootstrapped samples. WOC is consistently better than self-consistency and greedy decoding.}
\label{tab:results_elecpred}
\end{table*}

\section{Qualitative Examples of WOC Decoding Advantage}
\label{app:qual_examples}

To complement our quantitative results, we present qualitative examples that illustrate how WOC decoding can outperform self-consistency and greedy decoding. 

\paragraph{Example 1: Pathological Majority in the FUTURE Dataset.}  
Self-consistency decoding can fail when the most frequently sampled response is a clear outlier. For the question:

\textit{``In Q1 2023, Tesla’s total revenue in billions was 23.329. In Q1 2024, how many billions will Tesla’s total revenue be? Think step-by-step.''}

The ground truth is 21.3. However, \texttt{Mistral-7B} outputs ``0'' in 9 out of 30 samples due to reasoning that deems the question unanswerable. For example:

\begin{quote}
\small
\textit{``We don't have access to this information and it's too speculative to make an accurate guess. Therefore the final answer (arabic numerals) is 0.''}
\end{quote}

Other samples contain inflated estimates, such as:

\begin{quote}
\small
\textit{``Tesla's revenue growth rate has been increasing at a rate of approximately 50\% year-over-year... Final answer: 23.329 * 1.5\textasciicircum2 = 65.4231.''}
\end{quote}

In contrast, the response closest to the median was:

\begin{quote}
\small
\textit{``...Let's use the percentage growth rate from Q1 2022 to Q1 2023... Therefore the final answer (arabic numerals) is 25.681.''}
\end{quote}

Taking the median removes both extreme under- and over-estimates, resulting in a more stable and accurate prediction.

\paragraph{Example 2: The Lack of Majority in the MARBLES Dataset.}  
In some cases, self-consistency fails due to the lack of any majority answer. For the question:

\textit{``How many standard-sized U.S. marbles does it take to fill a one-cup dry ingredient measuring cup? Think step-by-step.''}

\texttt{Mixtral-8x7B} produced 30 unique answers ranging from 0 to 73,384, with a standard deviation of 17{,}654. For example:

\textbf{Extreme overestimate:}
\begin{quote}
\small
\textit{``...volume of a marble = 0.05236 in³, cup = 3843.88 in³ $\rightarrow$ 3843.88 / 0.05236 = 73384 marbles... Therefore the final answer (arabic numerals) is 73{,}384.''}
\end{quote}

\textbf{Extreme underestimate:}
\begin{quote}
\small
\textit{``...estimated size used is too small... it is safe to say that a one-cup dry measure cannot be filled with any standard-sized U.S. marbles. Therefore the final answer (arabic numerals) is 0 marbles.''}
\end{quote}

\textbf{Response near the median:}
\begin{quote}
\small
\textit{``...it would take approximately 70.6 standard-sized U.S. marbles to fill a one-cup dry ingredient measuring cup. Final answer: 71 marbles.''}
\end{quote}

The ground truth is 62 marbles. While self-consistency is unstable due to the lack of repetition, WOC decoding yields a median estimate of 81—much closer to the true answer.

\section{Predicted Election Outcomes Visualized on a National Map}
\label{app:elecpred_map}

\begin{figure*}[htb!]
    \centering
    \includegraphics[width=\linewidth]{figures/map_smalltext_all.pdf}
    \caption {Comparison of Actual and Predicted Vote Percentages in the 2024 U.S. Presidential Election (LLaMA-2-7b-Chat; ELECPRED dataset). (A) The actual vote percentage Kamala Harris received in each state in 2024 US presidential election. (B) The predicted vote percentage using wisdom of crowds (median) decoding. (C) The predicted vote percentage using greedy decoding. (D) The predicted vote percentage using self-consistency (majority) decoding. (E) The predicted vote percentage using mean decoding. For (B), (C), (D), and (E) the predicted vote percentage using each strategy is given, followed by the actual vote percentage in brackets.}
    \label{fig:election_pred_map}
\end{figure*}

As shown in  Table~\ref{tab:results_elecpred}, WOC decoding outperforms both self-consistency and greedy decoding in prediction accuracy in terms of the vote percentage Kamala Harris received in the 2024 U.S. presidential election. However, the difference in quality is difficult to interpret intuitively. To better illustrate the results, we visualize the predicted election outcomes on a national map (Figure~\ref{fig:election_pred_map}). While LLMs predicts the percentage of votes Kamala Harris would receive in each state, we convert these percentages into electoral votes to compare them with the actual election outcome, in which Donald Trump won 312 electoral votes, while Kamala Harris received 226. The results show that WOC decoding provides the closest prediction (194 electoral votes for Trump). In contrast, greedy decoding predicts 176, self-consistency predicts 148, and the mean baseline predicts 191. Notably, all greedy decoding, self-consistency, and mean decoding made implausible errors: greedy decoding predicts a Democratic win in Texas, self-consistency incorrectly predicts Democratic wins in Arkansas and Louisiana, and mean decoding predicts several states to have over 100\% of Democrat vote. While WOC decoding achieves the most accurate prediction, it shows an overall bias favoring Democrats. Understanding the source of this bias remains an open question for future research. 

\section{Statistical Tests}
\label{app:stat_tests}

For each dataset, we conducted a Wilcoxon signed-rank paired test \cite{wilcoxon1945} to test whether the WOC decoding is better than self-consistency and greedy decoding across LLMs.  Results show that WOC decoding tends to have smaller normalized error ($\varepsilon$) than self-consistency, MARBLES: $p < 0.001$, FUTURE: $p = 0.02$, ELECPRED: $p = 0.02$. When comparing WOC decoding and greedy decoding, WOC decoding is also significantly better than greedy decoding across LLMs, MARBLES: $p < 0.001$, FUTURE: $p = 0.02$, ELECPRED: $p = 0.01$.

\section{Distributional Robustness of WOC Decoding} 
\label{app:dist_analysis}

To investigate the mechanism behind the superiority of WOC decoding, we analyzed the distributional properties of model responses. Human WOC literature has shown that the median is a robust estimator regardless of the shape of the response distribution as long as their provided responses are statistically independent \citep{galton1907vox, surowiecki2005wisdom}. Inspired by this, we examined whether distributional properties like skewness or variance correlate with the effectiveness of WOC decoding in LLMs. The skewness of the response distribution varies greatly across LLMs ($[-1.19, 0.64]$ for ELECPRED, $[-1.60, 3.15]$ for FUTURE, and $[2.23, 2.56]$ for MARBLES). Furthermore, we found no consistent relationship between skewness and WOC effectiveness. Pearson's $r$ between skewness and WOC gain (i.e., the reduction in normalized error over self-consistency) also varies widely ($[-0.26, 0.64]$ for ELECPRED, $[-0.49, 0.27]$ for FUTURE, and $[-0.52, 0.41]$ for MARBLES). Similarly, standard deviation of the response distribution shows no clear correlation with WOC utility. These findings support the idea that WOC decoding is robust across diverse response distributions and the superiority is not tied to simple distribution properties.

\section{Selection of the LLMs}
\label{app:list_models}

Table ~\ref{tab:list_models} lists the LLMs that we evaluate. The knowledge cutoff dates were decided based on the model description webpage. For the Mistral and Mixtral models, the knowledge cutoff dates were not released, so the date listed is the date of model weight commits on HuggingFace \footnote{\url{https://huggingface.co/mistralai/Mistral-7B-Instruct-v0.2/commit/dca6e4b60aca009ed25ffa70c9bb65e46960a573}}\footnote{\url{https://huggingface.co/mistralai/Mixtral-8x7B-Instruct-v0.1/commit/858fdc292793fc3e671bf51fc5586c5cc10fbe3a}}\footnote{\url{https://huggingface.co/mistralai/Mixtral-8x22B-Instruct-v0.1/commit/796bc4393fd5e7e0c0ff1c44de2526419f163003}}.

\begin{table}[htbp!]

\centering
\small
\resizebox{\linewidth}{!}{%
\begin{tabular}{@{}p{2cm}p{4cm}p{3cm}@{}}
    \toprule
    Model Family & Model Variant & Knowledge Cutoff Date  \\        
    \midrule
    Mistral & mistral-7b-instruct-v0.2 & Before Dec. 2023 
    \\ 
    Mixtral & mixtral-8x7b-instruct-v0.1 & Before Dec. 2023 \\ 
    & mixtral-8x22b-instruct-v0.1 & Before Apr. 2024 \\ 
    LLaMA 2 & llama-2-7b-chat-hf & Jul. 2023 \\ 
    & llama-2-13b-chat-hf & Jul. 2023 \\ 
    & llama-2-70b-chat-hf & Jul. 2023 \\ 
    LLaMA 3.1 & llama-3.1-8b-instruct & Dec. 2023 \\ 
    & llama-3.1-70b-instruct & Dec. 2023 \\ 
    GPT & gpt-3.5-turbo-0125 & Sep. 2021 \\ 
    & gpt-4-0125-preview & Dec. 2023 \\ 

     \bottomrule

\end{tabular}
}
\caption{List of large language models.}

\label{tab:list_models}
\end{table}

\section{Guesstimation Questions and Ground Truth Answers}
\label{app:list_questions}

Tables ~\ref{tab:list_questions_marbles} and ~\ref{tab:list_questions_future} list the guesstimation questions used in the MARBLES and FUTURE datasets along with their corresponding ground truth answers. \\

The following sources were used to determine the ground truth answers for the FUTURE dataset:

\begin{itemize}
    \item\href{https://media.ford.com/content/dam/fordmedia/North%20America/US/2024/07/q2sales/Q2%202024%20Sales%20Final.pdf}{Ford Sales}

    \item\href{https://www.macrotrends.net/global-metrics/cities/23083/new-york-city/population}{New York City Population}

    \item\href{https://www.olympics.com/en/olympic-games/paris-2024/medals}{2024 Olympic Medal Table}, \href{https://www.olympics.com/en/olympic-games/tokyo-2020/medals}{2020 Olympic Medal Table}

    \item\href{https://fred.stlouisfed.org/series/GDP/}{United States GDP}

    \item\href{https://digitalassets.tesla.com/tesla-contents/image/upload/IR/TSLA-Q1-2024-Update.pdf}{Tesla Sales}

    \item\href{https://registrar.wisc.edu/enrollment-reports/}{University of Wisconsin-Madison Enrollment}

    \item\href{https://www.apple.com/newsroom/2024/02/apple-reports-first-quarter-results/}{Apple 2024 Sales}, \href{https://www.apple.com/newsroom/2023/02/apple-reports-first-quarter-results/}{Apple 2023 Sales}

    \item\href{https://www.njweather.org/content/scorching-june-2024-and-january%E2%80%93june-recaps}{New Jersey 2024 Temperature}, \href{https://www.njweather.org/monthly-summaries?page=1}{New Jersey 2023 Temperature}

    \item\href{https://www.sony.com/en/SonyInfo/IR/library/presen/er/pdf/24q1_supplement.pdf}{Sony Sales}

    \item\href{https://www.nytimes.com/2024/04/04/climate/global-forest-tree-loss-wri.html}{2023 Forest Loss}, \href{https://www.globalforestwatch.org/blog/forest-insights/global-tree-cover-loss-data-2022/}{2022 Forest Loss}

    \item\href{https://spacestatsonline.com/launches/year/2023}{2023 Satellite Launches}, \href{https://spacestatsonline.com/launches/year/2024}{2024 Satellite Launches}

    \item\href{https://fred.stlouisfed.org/series/ASPUS}{United States Home Prices}

    \item\href{https://fred.stlouisfed.org/series/ICSA#0}{United States Unemployment Claims}

    \item\href{https://www.tsa.gov/travel/passenger-volumes/2024}{2024 TSA Passenger Count}, \href{https://www.tsa.gov/travel/passenger-volumes/2023}{2023 TSA Passenger Count}

\end{itemize}

Table ~\ref{tab:list_state_election_results} lists the percentage of the vote Kamala Harris received in the 2024 presidential Election and number of electoral votes for each state and the District of Columbia. \\

The following is text is the format of the prompt for the ELECPRED dataset, where the results are listed for all presidential elections from 1976 to 2020: 

\begin{quote}
\tt \small Here is a history of prior voting results from the US state of Alabama for US Presidential elections:

1976:  Jimmy Carter (Democrat) versus  Gerald Ford (Republican). Carter (the Democrat) received 56 percent of the vote.

...

2020:  Joseph R. Jr Biden (Democrat) versus  Donald J. Trump (Republican). Biden (the Democrat) received 37 percent of the vote.

In the 2024 election, the candidates will be Vice President Kamala Harris (the Democrat) and former President Donald Trump (the Republican). What percentage of the vote in Alabama do you think Kamala Harris (the Democrat) will receive? You must not predict a tie.

\end{quote}

The historical results from each state can be found on the \href{https://history.house.gov/Institution/Election-Statistics/Election-Statistics/}{United States House of Representatives Archive} \citep{houseofrepselecresults}.

\begin{table*}[!htbp]
\centering
\small
\resizebox{\linewidth}{!}{
    \begin{tabular}{@{}p{15.5cm}l@{}}
        \toprule
        \textbf{Question} & \textbf{True Answer} \\
        \midrule
        How many standard-sized U.S. marbles does it take to fill a one cup dry ingredient measuring cup? & 62 \\
        How many standard-sized U.S. marbles does it take to fill a single-shot shot glass? & 13 \\
        How many standard-sized U.S. marbles does it take to fill a Starbucks iced tall cup? & 109 \\
        How many standard-sized U.S. marbles does it take to fill an Altoids tin container? & 22 \\
        How many standard-sized U.S. marbles does it take to fill the box for a deck of cards (standard-sized Bicycle playing cards)? & 24 \\
        How many standard-sized M\&Ms does it take to fill a one cup dry ingredient measuring cup? & 210 \\
        How many standard-sized M\&Ms does it take to fill a single-shot shot glass? & 51 \\
        How many standard-sized M\&Ms does it take to fill a Starbucks iced tall cup? & 382 \\
        How many standard-sized M\&Ms does it take to fill an Altoids tin container? & 95 \\
        How many standard-sized M\&Ms does it take to fill the box for a deck of cards (standard-sized Bicycle playing cards)? & 96 \\
        How many U.S. quarters does it take to fill a one cup dry ingredient measuring cup? & 160 \\
        How many U.S. quarters does it take to fill a single-shot shot glass? & 42 \\
        How many U.S. quarters does it take to fill a Starbucks iced tall cup? & 280 \\
        How many U.S. quarters does it take to fill an Altoids tin container? & 70 \\
        How many U.S. quarters does it take to fill the box for a deck of cards (standard-sized Bicycle playing cards)? & 70 \\
        \bottomrule
    \end{tabular}
    
}
\caption{List of all MARBLES questions and their corresponding true answers.}

\label{tab:list_questions_marbles}
\end{table*}

\begin{table*}[!htbp]
\centering
\small
\resizebox{\linewidth}{!}{
    \begin{tabular}{@{}p{15.5cm}l@{}}
        \toprule
        \textbf{Question} & \textbf{True Answer} \\
        \midrule
        In the second quarter of 2023, the number of vehicles Ford sold was 531662. In the second quarter of 2024, how many vehicles will Ford sell? & 536,050 \\
        In 2023 the population of the New York City Metropolitan Area was 18937000. In 2024, how many people will live in the New York City Metropolitan Area? & 19,034,000 \\
        In the 2020 Summer Olympics, the number of medals the United States won was 113. In the 2024 Summer Olympics, how many medals will the United States win? & 126 \\
        In Q2 2023, the United States' GDP in billions was 27453.815. In Q2 2024, how many billions will the United States' GDP be? & 29,016.714 \\
        In Q1 2023, Tesla's total revenue in billions was 23.329. In Q1 2024, how many billions will Tesla's total revenue be? & 21.301 \\
        In the 2023-24 school year, the number of students enrolled at the University of Wisconsin Madison was 50,633. In the 2024-25 school year, how many students will be enrolled at the University of Wisconsin Madison? & 52,097 \\
        In Q1 2023 Apple's total revenue in billions 117.2. In Q1 2024, how many billions will Apple's total revenue be? & 119.6 \\
        The average temperature in degrees Fahrenheit in New Jersey in June 2023 was 67.8. In June 2024, what will the average temperature in degrees Fahrenheit in New Jersey be? & 73.6 \\
        In Q1 2023 the number of PlayStation 5 units sold was 3300000. In Q1 2024, how many PlayStation 5 units will be sold? & 2,400,000 \\
        In Q1 2023 the number of monthly active users on the PlayStation Network in millions was 108. In Q1 2024, how many monthly active users in millions will the PlayStation Network have? & 116 \\
        In 2022 the number of acres of primary tropical forest lost was 10130000. In 2023, how many acres of primary tropical forest will be lost? & 9,100,000 \\
        The number of satellites the United States launched into space from January to October 2023 was 85. From January to October 2024, how many satellites will the United States launch into space? & 111 \\
        In Q1 2023 the average sale price of a house in the United States was 505300. In Q1 2024, what will the average sale price of a house in the United States be? & 519,700 \\
        In Q3 2023 the number of unemployment insurance claims filed was 232643. In Q3 2024, how many unemployment insurance claims will be filed? & 231,154 \\
        From January 2023 to the beginning of October 2023 the number of passengers that passed through TSA security in the United States was 638549095. From January 2024 to the beginning of October 2024, how many passengers will pass through TSA security in the United States? & 677,657,486 \\
        \bottomrule
    \end{tabular}
    
}
\caption{List of all FUTURE questions and their corresponding true answers.}

\label{tab:list_questions_future}
\end{table*}

\begin{table*}[!htbp]
\centering
\small
    \begin{tabular}{lll}
        \toprule
        \textbf{State} & \textbf{Electoral Vote Count} & \textbf{\% Harris Vote} \\
        \midrule
        Alabama & 9 & 34.1\% \\
        Alaska & 3 & 41.4\% \\
        Arizona & 11 & 46.7\% \\
        Arkansas & 6 & 33.5\% \\
        California & 54 & 58.6\% \\
        Colorado & 10 & 54.1\% \\
        Connecticut & 7 & 56.4\% \\
        Delaware & 3 & 56.6\% \\
        District Of Columbia & 3 & 90.3\% \\
        Florida & 30 & 43.0\% \\
        Georgia & 16 & 48.5\% \\
        Hawaii & 4 & 60.6\% \\
        Idaho & 4 & 30.4\% \\
        Illinois & 19 & 54.6\% \\
        Indiana & 11 & 39.6\% \\
        Iowa & 6 & 42.5\% \\
        Kansas & 6 & 41.0\% \\
        Kentucky & 8 & 33.9\% \\
        Louisiana & 8 & 38.2\% \\
        Maine & 4 & 52.1\% \\
        Maryland & 10 & 62.9\% \\
        Massachusetts & 11 & 60.9\% \\
        Michigan & 15 & 48.3\% \\
        Minnesota & 10 & 51.1\% \\
        Mississippi & 6 & 37.3\% \\
        Missouri & 10 & 40.0\% \\
        Montana & 4 & 38.3\% \\
        Nebraska & 5 & 39.1\% \\
        Nevada & 6 & 47.5\% \\
        New Hampshire & 4 & 50.7\% \\
        New Jersey & 14 & 51.8\% \\
        New Mexico & 5 & 51.9\% \\
        New York & 28 & 55.6\% \\
        North Carolina & 16 & 47.6\% \\
        North Dakota & 3 & 30.5\% \\
        Ohio & 17 & 43.9\% \\
        Oklahoma & 7 & 31.9\% \\
        Oregon & 8 & 55.3\% \\
        Pennsylvania & 19 & 48.6\% \\
        Rhode Island & 4 & 55.5\% \\
        South Carolina & 9 & 40.4\% \\
        South Dakota & 3 & 34.2\% \\
        Tennessee & 11 & 34.4\% \\
        Texas & 40 & 42.4\% \\
        Utah & 6 & 37.8\% \\
        Vermont & 3 & 63.8\% \\
        Virginia & 13 & 51.8\% \\
        Washington & 12 & 57.6\% \\
        West Virginia & 4 & 28.1\% \\
        Wisconsin & 10 & 48.8\% \\
        Wyoming & 3 & 25.8\% \\
        \bottomrule
    \end{tabular}
    
\caption{List of all state results for the United States 2024 presidential election.}

\label{tab:list_state_election_results}
\end{table*}

\section{The Prompts and Extraction used for querying the LLMs}

\label{app:list_prompts}

Table ~\ref{tab:list_prompts_marbles} lists the prompts that are used when querying the LLMs on the MARBLES dataset. Table ~\ref{tab:list_prompts_elecpred} lists the prompts that are used when querying the LLMs on the ELECPRED dataset. Table ~\ref{tab:list_prompts_future} lists the prompts that are used when querying the LLMs on the FUTURE dataset. Note the addition of the phrase "If you don't have enough information, just make a guess." to the FUTURE system prompts. We use a separate LLM (gpt-4-o-mini) to extract the numeric estimate from the CoT response. If the response can’t be parsed successfully, we resampled and parsed again up to 3 times. After this process, we have manually verified the extraction and the parsing accuracy is 100\%.

\begin{table*}[!htb]
\centering
\small
\resizebox{\linewidth}{!}{
    \begin{tabular}[t]{@{}llp{5.5cm}p{7cm}@{}}
        \toprule
        \textbf{Prompt Type} & \textbf{Message Type} & \textbf{Prompt} & \textbf{Example}\\
        \midrule
        Initial Prompt & \textit{System Message} & You must provide a final answer. & You must provide a final answer. \\
        \midrule
        Initial Prompt & \textit{User Message} & \makecell[tl]{\{question\} Think step-by-step.\\ You have to use the following format\\ Reasoning: [Your step-by-step reasoning]\\ Final answer: [A number. No other text or \\ explanation]} & \makecell[tl]{\{How many standard-sized M\&Ms does it take to fill \\ a Starbucks iced tall cup?\} Think step-by-step.
        You \\ have to use the following format \\
        Reasoning: [Your step-by-step reasoning] \\
        Final answer: [A number. No other text or explanation]}\\
        \midrule
        Two Step Extraction & \textit{User Message} & \{initial\_response\}. Therefore the final answer (arabic numerals) is & 
        \makecell[tl]{\{How many standard-sized M\&Ms does it take to fill \\ a Starbucks iced tall cup? Think step-by-step.
        You \\ have to use the following format \\
        Reasoning: [Your step-by-step reasoning] \\
        Final answer: [A number. No other text or explanation] \\ 
        Reasoning: \\
 A Starbucks iced tall cup has a volume of approximately \\ 12 oz or 355 ml. The volume of a single standard-sized \\ M\&M is estimated to be around 0.103 oz or 2.94 ml \\ based on the density of milk chocolate and average \\ dimensions of the candy. To calculate the number of \\ M\&Ms needed to fill the cup, we can convert the total \\ volume to M\&M volumes and round up to the nearest \\ M\&M to account for excess candy: \\
 Number of M\&Ms = Total volume / Volume of a single \\ M\&M \\
 Number of M\&Ms = 355 ml / 2.94 ml \\
 Number of M\&Ms = 121.63 = 122 M\&Ms \\
 Final answer: 122 M\&Ms.\} Therefore the final \\ answer (arabic numerals) is} \\
        \bottomrule
    \end{tabular}
}
\caption{The prompts used for query the LLMs on the MARBLES dataset.}

\label{tab:list_prompts_marbles}
\end{table*}
\begin{table*}[!htb]
\centering
\small
\resizebox{\linewidth}{!}{
    \begin{tabular}[t]{@{}llp{5.5cm}p{7cm}@{}}
        \toprule
        \textbf{Prompt Type} & \textbf{Message Type} & \textbf{Prompt} & \textbf{Example}\\
        \midrule
        Initial Prompt & \textit{System Message} & You must provide a final answer. If you don't have enough information, just make a guess. & You must provide a final answer. If you don't have enough information, just make a guess. \\
        \midrule
        Initial Prompt & \textit{User Message} & \makecell[tl]{\{question\} Think step-by-step.\\ You have to use the following format\\ Reasoning: [Your step-by-step reasoning]\\ Final answer: [A number. No other text or \\ explanation]} & \makecell[tl]{\{In the second quarter of 2023, the number of vehicles \\ Ford sold was 531662. In the second quarter of 2024, \\ how many vehicles will Ford sell?\} Think step-by-step. \\
        You have to use the following format \\
        Reasoning: [Your step-by-step reasoning] \\
        Final answer: [A number. No other text or explanation]}\\
        \midrule
        Two Step Extraction & \textit{User Message} & \{initial\_response\}. Therefore the final answer (arabic numerals) is & 
        \makecell[tl]{\{In the second quarter of 2023, the number of vehicles \\ Ford sold was 531662. In the second quarter of 2024, \\ how many vehicles will Ford sell? Think step-by-step. \\
        You have to use the following format \\
        Reasoning: [Your step-by-step reasoning] \\
        Final answer: [A number. No other text or explanation] \\ 
        Answer : 564250 \\ 
        Reasoning : \\
        The information given in the question is Second quarter \\ of 2023 - Ford sold 531662.\} Therefore the final \\ answer (arabic numerals) is} \\
        \bottomrule
    \end{tabular}
}
\caption{The prompts used for query the LLMs on the FUTURE dataset.}

\label{tab:list_prompts_future}
\end{table*}
\begin{table*}[!htbp]
\centering
\small
\resizebox{\linewidth}{!}{
    \begin{tabular}[t]{@{}p{2cm}p{2cm}p{3cm}p{10cm}@{}}
        \toprule
        \textbf{Prompt Type} & \textbf{Message Type} & \textbf{Prompt} & \textbf{Example}\\
        \midrule
        Initial Prompt & \textit{System Message} & You must provide a final answer. & You must provide a final answer. \\
        \midrule
        Initial Prompt & \textit{User Message} & \{question\} \newline Think step-by-step.\newline You have to use the following format\newline Reasoning: [Your step-by-step reasoning]\newline Final answer: [A number. No other text or explanation] & Here is a history of prior voting results from the US state of Alabama for US Presidential elections: \newline 
        1976:  Jimmy Carter (Democrat) versus  Gerald Ford (Republican). Carter (the Democrat) received 56 percent of the vote. \newline 
        1980:  Jimmy Carter (Democrat) versus  Ronald Reagan (Republican). Carter (the Democrat) received 49 percent of the vote. \newline 
        1984:  Walter Mondale (Democrat) versus  Ronald Reagan (Republican). Mondale (the Democrat) received 38 percent of the vote. \newline 
        1988:  Michael Dukakis (Democrat) versus  George H.W. Bush (Republican). Dukakis (the Democrat) received 40 percent of the vote. \newline 
        1992:  Bill Clinton (Democrat) versus  George H.W. Bush Republican). Clinton (the Democrat) received 46 percent of the vote. \newline 
        1996:  Bill Clinton (Democrat) versus  Robert Dole (Republican). Clinton (the Democrat) received 46 percent of the vote. \newline 
        2000:  Al Gore (Democrat) versus  George W. Bush (Republican). Gore (the Democrat) received 42 percent of the vote. \newline 
        2004:  John Kerry (Democrat) versus  George W. Bush (Republican). Kerry (the Democrat) received 37 percent of the vote. \newline 
        2008:  Barack H. Obama (Democrat) versus  John Mccain (Republican). Obama (the Democrat) received 39 percent of the vote. \newline 
        2012:  Barack H. Obama (Democrat) versus  Mitt Romney (Republican). Obama (the Democrat) received 38 percent of the vote. \newline 
        2016:  Hillary Clinton (Democrat) versus  Donald J. Trump (Republican). Clinton (the Democrat) received 35 percent of the vote. \newline 
        2020:  Joseph R. Jr Biden (Democrat) versus  Donald J. Trump (Republican). Biden (the Democrat) received 37 percent of the vote. \newline 
        In the 2024 election, the candidates will be Vice President Kamala Harris (the Democrat) and former President Donald Trump (the Republican). What percentage of the vote in Alabama do you think Kamala Harris (the Democrat) will receive? You must not predict a tie.\} \newline 
        Think step-by-step. \newline 
        You have to use the following format \newline 
        Reasoning: [Your step-by-step reasoning] \newline 
        Final answer: [A number. No other text or explanation] 
        \\
        \midrule
        Two Step Extraction & \textit{User Message} & \{initial\_response\}. Therefore the final answer (arabic numerals) is & 
        
        \{Here is a history of prior voting results from the US state of Alabama for US Presidential elections: \newline 
        1976:  Jimmy Carter (Democrat) versus  Gerald Ford (Republican). Carter (the Democrat) received 56 percent of the vote. \newline 
        1980:  Jimmy Carter (Democrat) versus  Ronald Reagan (Republican). Carter (the Democrat) received 49 percent of the vote. \newline 
        1984:  Walter Mondale (Democrat) versus  Ronald Reagan (Republican). Mondale (the Democrat) received 38 percent of the vote. \newline 
        1988:  Michael Dukakis (Democrat) versus  George H.W. Bush (Republican). Dukakis (the Democrat) received 40 percent of the vote. \newline 
        1992:  Bill Clinton (Democrat) versus  George H.W. Bush (Republican). Clinton (the Democrat) received 46 percent of the vote. \newline 
        1996:  Bill Clinton (Democrat) versus  Robert Dole (Republican). Clinton (the Democrat) received 46 percent of the vote. \newline 
        2000:  Al Gore (Democrat) versus  George W. Bush (Republican). Gore (the Democrat) received 42 percent of the vote. \newline 
        2004:  John Kerry (Democrat) versus  George W. Bush (Republican). Kerry (the Democrat) received 37 percent of the vote. \newline 
        2008:  Barack H. Obama (Democrat) versus  John Mccain (Republican). Obama (the Democrat) received 39 percent of the vote. \newline 
        2012:  Barack H. Obama (Democrat) versus  Mitt Romney (Republican). Obama (the Democrat) received 38 percent of the vote. \newline 
        2016:  Hillary Clinton (Democrat) versus  Donald J. Trump (Republican). Clinton (the Democrat) received 35 percent of the vote. \newline 
        2020:  Joseph R. Jr Biden (Democrat) versus  Donald J. Trump (Republican). Biden (the Democrat) received 37 percent of the vote. \newline 
        In the 2024 election, the candidates will be Vice President Kamala Harris (the Democrat) and former President Donald Trump (the Republican). What percentage of the vote in Alabama do you think Kamala Harris (the Democrat) will receive? You must not predict a tie. \newline 
        Think step-by-step. \newline 
        You have to use the following format \newline 
        Reasoning: [Your step-by-step reasoning] \newline 
        Final answer: [A number. No other text or explanation] \newline 
        Reasoning:  Alabama has consistently voted for the Republican candidate in US Presidential elections. The state has a voter population of 3,894,973.\} Therefore the final answer (arabic numerals) is \\
        \bottomrule
    \end{tabular}
}
\caption{The prompts used for query the LLMs on the ELECPRED dataset.}

\label{tab:list_prompts_elecpred}
\end{table*}

\section{Human Experiment}
\label{app:human_exp}
To replicate previous findings about WOC in human crowds, and compare the LLMs' guesstimation performance with humans,
we recruited 230 participants from a U.S. university. Participants received course credit for their participation. Each participant was asked to provide estimates for each question in the MARBLES dataset. We also asked participants to rate their familiarity with each item and container on a 5-point scale (from 1 = ``not familiar at all'' to 5 = ``extremely familiar''). For each question, we only used data from participants who rated their familiarity as at least 4 (``quite familiar'') for both the item and the container, yielding an average of 64.9 valid responses per question. We conducted a human experiment only for the MARBLES dataset to ensure genuine guesstimation without easy access to the ground truth, as participants might already know the answers to some questions in the FUTURE and ELECPRED datasets.

\section{German Federal Election Prediction Experiment}

\label{app:result_german_elecpred}

To examine the cultural generalizability of our findings, we conducted a supplementary experiment on the 2025 German Federal Election. This experiment has similar structure as the ELECPRED dataset. Specifically, we asked LLMs to predict the vote share that the Social Democratic Party of Germany (SPD) would receive in the 2025 election across all 16 federal states.

\paragraph{Dataset Construction.} For each state, we provided historical vote percentages for SPD from elections spanning 1980 to 2021 \footnote{\url{https://github.com/awiedem/german_election_data/tree/main}}. We then asked the LLMs to predict the SPD's percentage share in the 2025 election. All prompts used the same chain-of-thought and sampling methodology as the ELECPRED setup.

\paragraph{Results.} The results show that WOC decoding does not consistently outperform other decoding strategies in this non-U.S. context. Table~\ref{tab:results_german_elecpred} presents the normalized errors across decoding strategies for each state. While WOC decoding remains competitive in some cases, its advantage is less pronounced than in ELECPRED. In addition, the normalized error $\varepsilon$ also tends to be larger than in ELECPRED. This observation is consistent with prior findings that LLMs are typically more accurate in U.S.-focused domains due to training data biases \citep{chu2024fairness}.

\begin{table}[!thbp]
\centering
\Large
\resizebox{\linewidth}{!}{
\begin{tabular}{l l l l}
\toprule
\textbf{Model} & \textbf{Wisdom of Crowds} & \textbf{Self-Consistency} & \textbf{Greedy} \\
& (WOC; Median) & (Majority) & \\
\midrule
Mistral & & & \\
\hspace{1em} mistral-7b-instruct-v0.2 & 0.61 [0.53, 0.69] & 0.62 [0.54, 0.70] & \textbf{0.50} [0.40, 0.60] \\
Mixtral & & & \\
\hspace{1em} mixtral-8x7b-instruct-v0.1 & 0.64 [0.54, 0.74] & 0.68 [0.57, 0.80] & \textbf{0.60} [0.48, 0.71] \\
LLaMA 2 & & & \\
\hspace{1em} llama-2-7b-chat & 1.11 [0.96, 1.25] & 1.14 [0.98, 1.29] & \textbf{1.09} [0.95, 1.23] \\
\hspace{1em} llama-2-70b-chat & \textbf{0.71} [0.61, 0.81] & 0.73 [0.63, 0.84] & 0.77 [0.65, 0.89] \\
LLaMA 3 & & & \\
\hspace{1em} llama-3.1-8b-instruct & 0.55 [0.48, 0.63] & \textbf{0.53} [0.44, 0.62] & 0.60 [0.47, 0.73] \\
\hspace{1em} llama-3.1-70b-instruct & \textbf{0.68} [0.55, 0.80] & 0.67 [0.54, 0.80] & 0.72 [0.59, 0.86] \\
GPT & & & \\
\hspace{1em} gpt-3.5-turbo-0125 & 0.61 [0.51, 0.71] & \textbf{0.51} [0.41, 0.61] & 0.65 [0.57, 0.73] \\
\hspace{1em} gpt-4-0125-preview & \textbf{0.96} [0.92, 1.00] & 1.00 [1.00, 1.00] & 0.83 [0.77, 0.90] \\
\bottomrule
\end{tabular}
}
\caption{Normalized errors ($\varepsilon$) for LLMs on the 2025 German Federal Election prediction task, following the same format as ELECPRED. Brackets indicate 95\% confidence intervals based on 30 bootstrapped samples. While WOC decoding remains competitive, its benefit is less consistent than in the U.S.-based ELECPRED dataset.}
\label{tab:results_german_elecpred}
\end{table}

\paragraph{Conclusion.}  Future work should explore guesstimation performance across diverse cultural contexts to evaluate the generalizability of LLM's guesstimation ability and the WOC strategy.

\section{Compute Resources}
\label{app:compute_resources}
We ran all experiments on a GPU machine equipped with 2x NVIDIA A100.

\end{document}